\documentclass{article}

\usepackage{arxiv}

\usepackage[utf8]{inputenc} 
\usepackage[T1]{fontenc}    
\usepackage{hyperref}       
\usepackage{url}            
\usepackage{booktabs}       
\usepackage{amsfonts}       
\usepackage{nicefrac}       
\usepackage{microtype}      
\usepackage{lipsum}
\usepackage{graphicx}
\usepackage{amsmath}
\usepackage{cleveref}
\usepackage{pslatex}
\usepackage{subcaption}
\graphicspath{ {./images/} }

\title{Combined Model for Partially-Observable and Non-Observable Task Switching: Solving Hierarchical Reinforcement Learning Problems Statically and Dynamically with Transfer Learning}

\author{
  Nibraas Khan \\
  Department of Computer Science\\
  Middle Tennessee State University\\
  Murfreesboro, TN 37132 \\
  \texttt{nak2z@mtmail.mtsu.edu} \\
   \And
  Joshua Phillips \\
  Department of Computer Science\\
  Middle Tennessee State University\\
  Murfreesboro, TN 37132 \\
  \texttt{Joshua.Phillips@mtsu.edu} \\
}

\begin{document}
\maketitle
\begin{abstract}
An integral function of fully autonomous robots and humans is the ability to focus attention on a few relevant percepts to reach a certain goal while disregarding irrelevant percepts. Humans and animals rely on the interactions between the Pre-Frontal Cortex (PFC) and the Basal Ganglia (BG) to achieve this focus called Working Memory (WM). The Working Memory Toolkit (WMtk) was developed based on a computational neuroscience model of this phenomenon with Temporal Difference (TD) Learning for autonomous systems. Recent adaptations of the toolkit either utilize Abstract Task Representations (ATRs) to solve Non-Observable (NO) tasks or storage of past input features to solve Partially-Observable (PO) tasks, but not both. We propose a new model, PONOWMtk, which combines both approaches, ATRs and input storage, with a static or dynamic number of ATRs. The results of our experiments show that PONOWMtk performs effectively for tasks that exhibit PO, NO, or both properties.
\end{abstract}


\section{Introduction}
In the pursuit of autonomous systems that mimic living beings, certain fundamental abilities are required. For a system to be autonomous, sensing, perception, cognition, planning, control, and actuation are integral \cite{Fukuda_how_far_away_is_man}. As a result, there have been many attempts to address the problem of perceptual learning (the ability to form representations of sensory information based on statistical information at the perceptual level).

Many autonomous systems do not perceive the world in the same way that humans do. Thus, there is no way to guarantee that autonomous systems can work well in environments where objects are defined by human design. So the systems must have the ability to form internal representations on their own \cite{Tugcu_neuroscience_model_of_working_memory}.

In a realistic environment, systems will be presented with large numbers of sensory stimuli, some irrelevant. A noted way of achieving focus is through the Working Memory Toolkit (WMtk), in which focusing on relevant percepts for the completion of a task results in a reward \cite{jovanovich_n_task,phillips_wmtk,dubois_hrr} - thus modeling Pre-Frontal Cortex (PFC) - Basal Ganglia (BG) interactions for Working Memory (WM) with a reward system \cite{Reilly_neuromodulatory,collins_neurogenetic}.

The WMtk was created for easy integration of a neural network-based WM model within autonomous systems by mitigating complex, internal details. With the use of the toolkit, autonomous systems can exhibit several key WM functions: focus on relevant task details, limit the search space through reward-based learning, and behave robustly \cite{baddeley_working_memory}.

Reinforcement Learning (RL) algorithms, specifically Temporal Difference (TD) Learning algorithms, work well when the Markov property is met \cite{kunz_td}. The original WMtk is successful when the Markov property is not met, specifically for Partially-Observable (PO) tasks \cite{Reilly_neuromodulatory,phillips_wmtk,dubois_hrr} such as a game of poker. The WMtk essentially turns a Non-Markovian (NM) task into a Markovian task by using WM. However, it struggles in situations when the environment provides no relevant information at any time (Non-Observable) such as the Wisconsin Card Sorting Test. The toolkit is ineffective when several sequential, conflicting tasks need to be learned which are NM and Non-Observable (NO). 

The solution to this NM-NO problem is an adaptation of the WMtk known as the n-task learning algorithm (nTL), which serves as an extension to the TD Learning framework \cite{jovanovich_n_task}. The algorithm works by forming ATRs based on reward feedback as opposed to perceptual features. The model uses ATRs, which are analogous to lenses that can observe the environment, directing attention across different subsets of features within a common state space.

There are two distinct approaches to focusing attention on relevant percepts: WM based on gating in relevant information and ATRs based on different understandings of the same environment. The two models address distinct problems in which the Markov property is not met, PO and NO, respectively. In this work, we propose a new model, which combines both approaches to solve complex tasks with both PO and NO components called the PONOWMtk to solve tasks such as chess with different strategies.

\section{Background}
\subsection{Working Memory and the Working Memory Toolkit}
Computational neuroscience defines WM in terms of the interactions between the PFC and the BG as observed in primates. A model for the interactions is TD learning, where the learning of relevant information about stimuli or actions is based on the rewards and punishments associated with them \cite{sutton_RL_intro}. In a single-layer neural network implementation, the value function (sum of discounted future rewards) is calculated using a simple dot product between a stimulus vector $\vec{u}$ and a weight vector $\vec{w}$:
\begin{equation}
    v_t = \vec{w}^T\vec{u}_t
\end{equation} 

A learning rule is used to update the weight vector, but first, the error, $\delta$, must be calculated:
\begin{equation}
    \delta_t = (r_t + \gamma \cdot v_{t+1}) - v_t
\end{equation} where $v_t$ is the predicted sum of future rewards at the current time step, $v_{t+1}$ is the estimated sum of future rewards at the next time step $t+1$, and $\gamma$ is the reward discount factor. 

With the error, the weights can be updated using a Rescorla-Wagner-like rule defined as:
\begin{equation}
    \vec{w}_{t+1} \xleftarrow{} \vec{w}_t + \alpha \cdot \delta_t \cdot \vec{u}_t 
\end{equation} where $\alpha$ is the learning rate. As the model learns, the value function converges to the actual sum of discounted rewards.

For WM, the RL framework is modified so that $\vec{u}$ consists of a conjunction of both the current perceptual features and potential features for storage in neural circuits analogous to the PFC. Due to the importance of WM, Noelle and Phillips created the original
set of software tools for developing working memory systems that can be easily integrated into robotic control mechanisms known as the WMtk \cite{phillips_wmtk}. The toolkit consists of a set of classes and methods that allows for the construction of a WM system that
uses TD learning to choose working memory content. The original toolkit works through the aid of a neural network for decisions about memory management, configurable parameters, user-defined reward functions, and user-defined release of useless WM \cite{Tugcu_neuroscience_model_of_working_memory}.

\subsection{Holographic Reduced Representations}

The original toolkit mitigates many challenges of integrating WM into a learning system but fails to provide aid to the user for the development of reasonable representations of the environment and WM concepts. The toolkit uses a neural network for learning, so these representations and concepts need to be encoded using a sparse, distributed formalism. It is difficult, even for experts, to develop and implement good representations. Evan for a simple binary encoding of two WM concepts, the user must define a function that produces a two-element vector encoding \cite{dubois_hrr}. By forcing the user to manually create functions such as these, the toolkit is prone to errors that can be mitigated by an automatic encoding process, and the toolkit cannot adapt to varying WM demands. 

To solve the problem of automatic encoding, the toolkit was integrated with a Holographic Reduced Representation Engine (HRRE). The purpose of the engine is to provide all the necessary capabilities to solve the automatic symbolic encoding (SE) to distributed encoding (DE) conversion. In the HRR formalism, independent representations are defined by a distributed vector of real numbers \cite{plate_holographic}. The engine can generate a DE based on a SE represented by a string.

Individual representations can be combined and reduced to a single vector that represents the combined knowledge of its constituents through a mathematical operation known as circular convolution. The combined representation retains the knowledge of both its constituents while the length of the combined vector and the constituents remains the same. Additionally, HRRs form a sparse, distributed formalism for compatibility with the WMtk’s underlying neural network architecture, allowing the same neural network to process increasingly complex concepts without modification to the architecture. Each representation is tied to a unique vector representation, DE, so each HRR can be tied to a complementary SE representation.

At its core, HRRs are vectors of real numbers that are typically drawn from a Normal distribution with zero mean $(\mu=0)$, and standard deviation, $\sigma = \frac{1}{\sqrt{n}}$ where $n$ is the length of the vectors. Orthogonality, near-zero dot product, between all HRRs and all convolutions of HRRs allows for robust learning of the function, $v$.

\subsection{Abstract Task Representations}

The HWMtk is contingent on the presence of a reward predicting stimulus at some time during the PO task. However, Policy changes that are driven by NO information can lead to the model failing when learning several conflicting tasks sequentially. A solution to this problem is found in nTL. nTL allows for any member of the TD learning family of algorithms to better handle scenarios in which the agent is required to switch between several tasks with different optimal policies. nTL uses ATRs to identify and separate tasks by using the feedback from the critic - TD error, $\delta$, in particular.

nTL can be used as an extension to any TD learning algorithm, but we arbitrarily use SARSA below. The action selection equation with the nTL extension becomes:

\begin{equation}
    m = \operatorname*{argmax}_{c \ \in \ C} ((\vec{s} \wedge \vec{c} \wedge \vec{atr}) \cdot \vec{w}_q + b)
\end{equation} where $\wedge$ is circular convolution, $m$ is the action chosen, $s$ is the current state representation, $C$ is the set of all candidate action choices for the current trial, $atr$ is the current representation in memory, $w_q$ is the weight vector for the $Q$ function neural network, and $b$ is the scalar bias term. 

The weight update becomes:
\begin{equation}
    \Delta w_i = \alpha_q[\mathrm{sgn}(\delta) \cdot \log(|\delta| + 1) \cdot (\vec{s} \wedge \vec{m} \wedge \vec{atr})_i]
\end{equation} where $w_i$ is the value of the weight vector at index $i$, $\alpha_q$ is the learning rate, $\delta$ is the error, and $(\vec{s} \wedge \vec{m} \wedge \vec{atr})_i$ is the HRR input vector at index $i$.

Each ATR is associated with an independent value function as well, which is updated with the TD error. In the equation below, $A$ is the function determining the ATR values, $\alpha_a$ is the learning rate for the ATRs, and $\delta$ is $r-A(\vec{atr})$:
\begin{equation}
    A(atr) \leftarrow A(\vec{atr}) + \alpha_a[\mathrm{sgn}(\delta) \cdot \log(|\delta| + 1)]
\end{equation}

When the TD error crosses a threshold, $t$, the model substitutes the next ATR in sequential order. The $t$ value is first set to negative one times the reward for the goal state, and is then updated at each time step using the TD error from the $Q$ function, where $\alpha_{at}$ is the learning rate:
\begin{equation}
    t \leftarrow t - \alpha_{at}[\mathrm{sgn}(\delta) \cdot \log(|\delta| + 1)]
\end{equation} 

The $t$ value is not updated for a task change external to the agent. nTL is made to work for both a preset static number of tasks and a dynamic number of tasks \cite{jovanovich_n_task}.

We hypothesize that a synthesis of the HWMtk and nTL would provide a theoretical framework that is capable of solving real world problems with PO and NO features. However, it is not clear whether additional mechanisms must be developed to form this synthesis. In particular, nTL only triggers ATR swaps for large, unanticipated negative values of $\delta$, but PO problems sometimes exhibit large, unanticipated positive $\delta$ values as well. If not carefully considered, the wrong ATR may be rewarded with a positive $\delta$ and result in learning instabilities. Therefore, we also anticipate that a mechanism for handling large, unanticipated positive $\delta$s will be needed.

\section{Methods}

\subsection{Model Description}

At every time step, the agent needs to take into account the state, signal, WM, ATR, and reward with: 
\begin{equation}
    \vec{u} = (\vec{s} \wedge \vec{p} \wedge \vec{wm} \wedge \vec{atr} \wedge \vec{r})
\end{equation} where $\vec{u}$ is the HRR representation of all the relevant information (all vectors are HHRs): $\vec{s}$ is the state, $\vec{p}$ is the signal, $\vec{wm}$ is the WM, $\vec{atr}$ is the ATR, and $\vec{r}$ is the reward. 

The possible values for WM are the internal representation of the signal, previous WM, or the identity HRR. There will always be a vector present for the ATR so that the agent is never left without context. The signal is a PO feature that is only available for the first time step and is just an identity HRR for the rest of the time steps. The reward is only present when there is a goal at the state the agent is in, otherwise, it is just an identity HRR.

With the representation, $\vec{u}$, its value needs to be calculated for the agent to make appropriate decisions with a simple one-layer neural network where the weights are initialized as a HRR vector and the bias $(b)$ is set to one (optimistic critic). The value, $v$, is defined as: 
\begin{equation}
    v(\vec{u}) = (\vec{u} \cdot \vec{w}) + b
\end{equation} where $\cdot$ is the dot product and $\vec{w}$ is the weights of the network.

To update the weights, TD error, $\delta$, needs to be calculated using: 
\begin{equation}
    \delta_t = (r_t - \gamma \cdot \mathrm{v}(\vec{u}_{t+1})) - \mathrm{v}(\vec{u}_t)
\end{equation} where $r_t$ is the scalar reward value. 

An eligibility trace allows for a backward view of the steps as opposed to the usual forward view for more stable learning. On each time step, the trace is scaled using $\lambda$ for all previous states. The eligibility trace in terms of time, $t$, is defined as:

\begin{equation}
    \vec{e}_t = \lambda \cdot \vec{e}_{t-1} + \vec{u}_t
\end{equation} where $\vec{e}_{t-1}$ is the discounted accumulation of all other previous states, and $\vec{u}_t$ is the current state.

The weight update at time $t$ for the neural network can now be defined as: 
\begin{equation}
    \vec{w}_t = \vec{w}_{t-1} + \alpha \cdot \mathrm{logmod}(\delta_t) \cdot \vec{e}_t
\end{equation}

In the above equation, $\vec{w}_{t-1}$, is the weight vector at the previous time step, $\alpha$ is the learning rate, $logmod$ is a log-modulus transform (to stabilize learning by scaling error), and $\vec{e}_t$ is the eligibility trace at the current time step. 

At every time step, the agent must decide what move to make. The move is based on the value of the potential states the agent can step into. The maximum value of the next state and WM can be calculated using: 

\begin{equation}
    m, c = \underset{\vec{s} \ \in \ S, \ \vec{wm} \ \in \ WM}{argmax}(\mathrm{v}(\vec{s} \wedge \vec{p} \wedge \vec{wm} \wedge \vec{atr} \wedge \vec{r}))
\end{equation}

The above equation uses the $argmax$ function where the agent enumerates through states, $\vec{s}$, of all possible states $S$ and all working memory content, $\vec{wm}$, of all possible working memory contents $WM$. At any time step $t$, the agent can use the above equation to decide the move, $m$ (external decision) and WM, $c$ (internal decision).

In case the agent is stuck in a local minimum, we implement an epsilon soft policy, $\epsilon$, which allows for the agent to make exploratory decisions some small fraction of time steps, $\epsilon$. During the random move, $c$ and the $\vec{atr}$ are not affected, only $m$ is.

The agent must also decide which context to use independent of the decision above. The context switch is triggered by the error rather than maximum value estimates. When $\delta_t$ crosses a certain threshold, the context switch is triggered. The threshold, $t$, in the model can be either a static or dynamic hyper-parameter. With the dynamic threshold, a threshold alpha, $\alpha_t$, is used to update $t$ at every time step (where $t$ initially set to one):
\begin{equation}
    t \mathrel{{+}{=}} \alpha_t \cdot \mathrm{logmod}(\delta)
\end{equation}

When a large negative error occurs that crosses $-t$, the agent interprets this error to mean that the wrong $atr$ was used (i.e. the task demands must have changed and requires a different ATR). For this case, the agent chooses the next $atr$ sequentially rather than choosing the highest value because the correct $atr$ cannot be determined directly for the case of large negative errors.

\begin{table*}[t]
\centering
\small
\setlength\tabcolsep{5pt}
\begin{tabular}{c c c c c c c} 
 \textbf{Parameters} & \textbf{Static PO} & \textbf{Dynamic PO} & \textbf{Static NO} & \textbf{Dynamic NO} & \textbf{Static PONO} & \textbf{Dynamic PONO} \\
 \hline
 \hline
 Hrr Length $(n)$ & 10240 & 15360 & 6144 & 7168 & 25600 & 25600 \\
 Discount $(\gamma)$ & 0.9 & 0.9 & 0.7 & 0.7 & 0.7 & 0.7 \\
 Alpha for Training $(\alpha_{train})$ & 0.3 & 0.3 & 0.1 & 0.1 & 0.1 & 0.1 \\
 Alpha for Testing $(\alpha_{test})$ & 0.01 & 0.01 & 0.01 & 0.01 & 0.01 & 0.01 \\
 Epsilon Soft $(\epsilon_{test})$ & 0.00001 & 0.00001 & 0.00001 & 0.00001 & 0.00001 & 0.00001 \\
 Threshold $(t)$ & 0.3 & 1 & 0.3 & 1 & 0.3 & 1 \\
 Discount for the trace $(\lambda)$  & 0.0 & 0.0 & 0.0 & 0.0 & 0.01 & 0.01 \\
 ATR Alpha $(\alpha_a)$ & - & - & 0.00065 & 0.00063 & 0.00011 & 0.00011 \\
 ATR Threshold $(\alpha_t)$ & - & - & -0.5 & -0.5 & -0.5 & -0.5 \\
 Signals $(\vec{p})$ & R, G, B & R, G, B & - & - & R, G & R, G \\ 
 Goals & 3, 10, 14 & 3, 10, 14 & 0, 4, 7, 10, 13 & 0, 4, 7, 10, 13 & 2, 5; 8, 13 & 2, 5; 8, 13 \\
 Episodes & 100000 & 100000 & 100000 & 100000 & 100000 & 100000 \\ 
 Max Steps Per Episode & 100 & 100 & 100 & 100 & 100 & 100 \\
 NO Task Switch Rate & 500 & 500 & 500 & 500 & 1000 & 1000 \\ 
 
 \hline
\end{tabular}
\caption{Major parameters for all the models are shown in this table, but all parameters can be found on the GitHub page. These parameters are the same for both the reset and transfer method.}
\label{params}
\end{table*}

When the positive, $t$, is crossed with a large positive TD error, a $argmax$ function is used to determine the next $atr$: 
\begin{equation}
    atr \leftarrow \underset{\vec{atr} \ \in \ ATR}{argmax}(\vec{s} \wedge \vec{p} \wedge \vec{wm} \wedge \vec{atr} \wedge \vec{r})
\end{equation}

When the agent switches $atrs$ (for either large positive or large negative errors), $e_t$ is cleared out so that the agent does not learn under the wrong context. 

PONOWMtk can be used for an arbitrary number of tasks, $n$, so the model needs to have the ability to grow along with the task. Without growth, the HRRs and the neural network might not be able keep up with the required orthogonality. To facilitate growth, at every time step where the threshold is crossed, both positive and negative, and the mean of the ATR values crosses $\alpha_t$, the ATR values are reset, the threshold for task switching is reset (for dynamic threshold to one), the HRR size is increased linearly, the weights are reset, and the eligibility trace is reset.  

With this method, which we call the Reset Method (RM), all of the HRRs including the WMs, ATRs, and all internal representations will be lost, and the learned value function is also lost. However, and alternate method, called Transfer Method (TM), allows the model to transfer the values from the old, smaller neural network to the larger version. When the model grows, new HRRs are created for the internal representations, the pseudo-inverse of the new HRRs is calculated, and the new weights are calculated using matrix multiplication between the inverses and the values of the old HRRs. 

The HRR length, $n$, is updated using:
\begin{equation}
    n \leftarrow \frac{atr_c \cdot n}{atr_c - 1}
\end{equation} where $atr_c$ is the current number of $atrs$.

The new weights, $\vec{w}$, are set with:
\begin{equation}
    \vec{w} \leftarrow [\underset{\vec{u}_{new} \ \in \ U_{new}}{\vec{u}_{new}}]^{-1} * [\underset{\vec{u}_{old} \ \in \ U_{old}}{\mathrm{v}}(\vec{u}_{old})]
\end{equation} where $[\underset{\vec{u}_{new} \ \in \ U_{new}}{\vec{u}_{new}}]^{-1}$ is the inverse of the new, empty stacked HRRs, $[\underset{\vec{u}_{old} \ \in \ U_{old}}{\mathrm{v}}(\vec{u}_{old})]$ is the vector of the previous HRR values, and $*$ is the matrix multiplication. 

The TM is computationally efficient but becomes increasingly expensive with more tasks. At some point, using the RM might become more computationally tractable but this is beyond the scope of our current work.

The model can keep old knowledge (the partially learned value function), but also have the ability to learn new information without losing orthogonality. The new HRRs have the same values as the old HRRs and all the mathematical and constituent properties are preserved.

\subsection{Test Protocols}
We provide three tasks that test the effectiveness of our model. The three tasks test the PO, NO, and PONO features of the PONOWMtk. Within these tasks, we compare the effectiveness of the RM and TM. 

For all three tasks, the accuracy (whether the optimal number of steps were taken including ATR switching) was calculated for the last $10\%$ of the episodes. To make sure that the testing was stable, $\alpha_{test}$ was set to $0.01$ and $\epsilon_{text}$ was set to $0$. Additional experimental parameters are shown in Table \ref{params}.

Each of the three tasks, PO, NO, and PONO, were tested using both static and dynamic thresholds with a hundred seeds for both methods using GNU Parallel \cite{Tange_parallel}.

\begin{figure*}[t]
\centering
    \begin{subfigure}[b]{0.33\textwidth}
    \centering
            \includegraphics[width=\textwidth,clip=false]{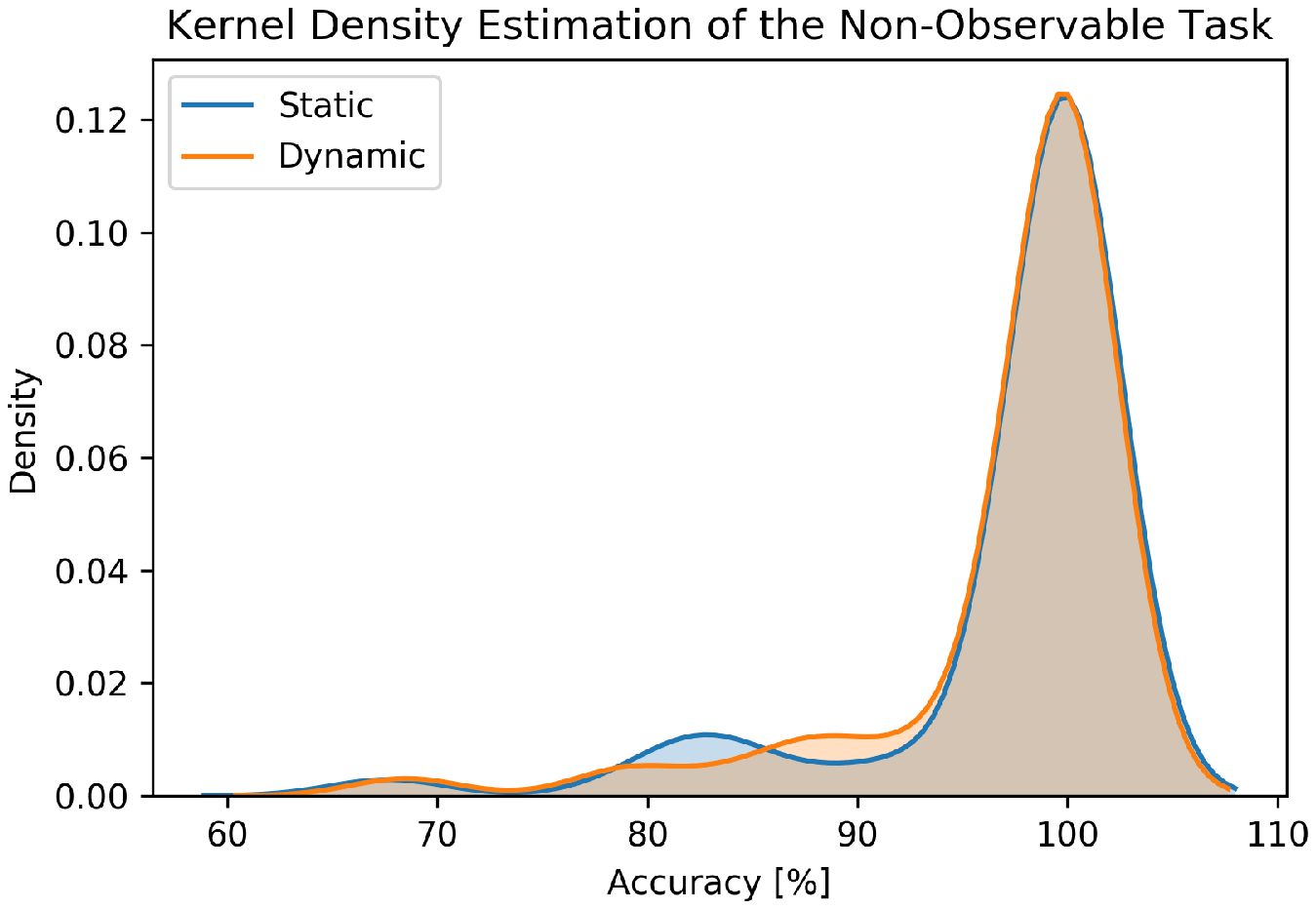}
            \caption{NO task with reset method}
            \label{reset_non_obs}
    \end{subfigure}
    \begin{subfigure}[b]{0.33\textwidth}
    \centering
            \includegraphics[width=\textwidth,clip=false]{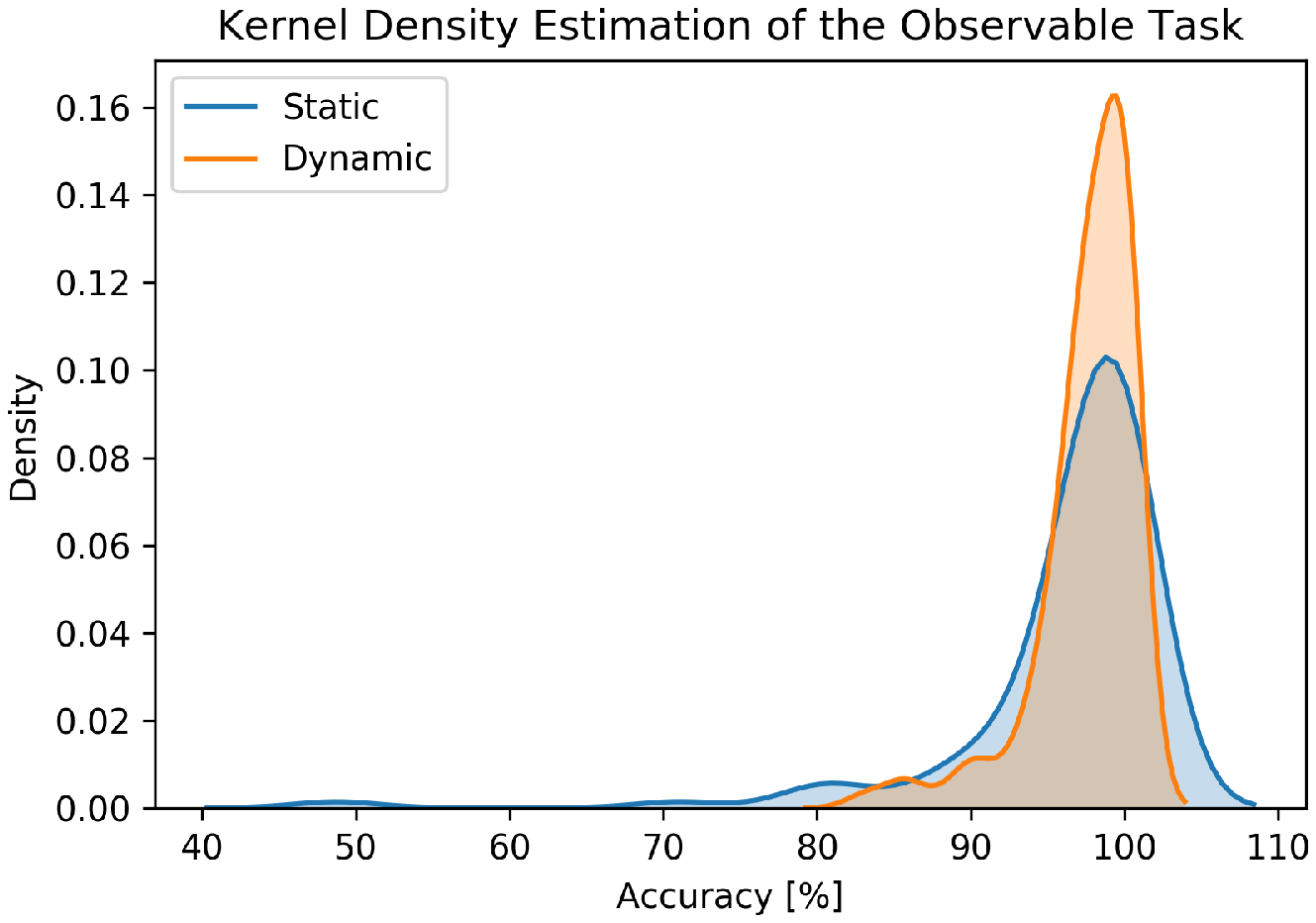}
            \caption{PO task with reset method}
            \label{reset_obs}
    \end{subfigure}
    \begin{subfigure}[b]{0.33\textwidth}
    \centering
            \includegraphics[width=\textwidth,clip=false]{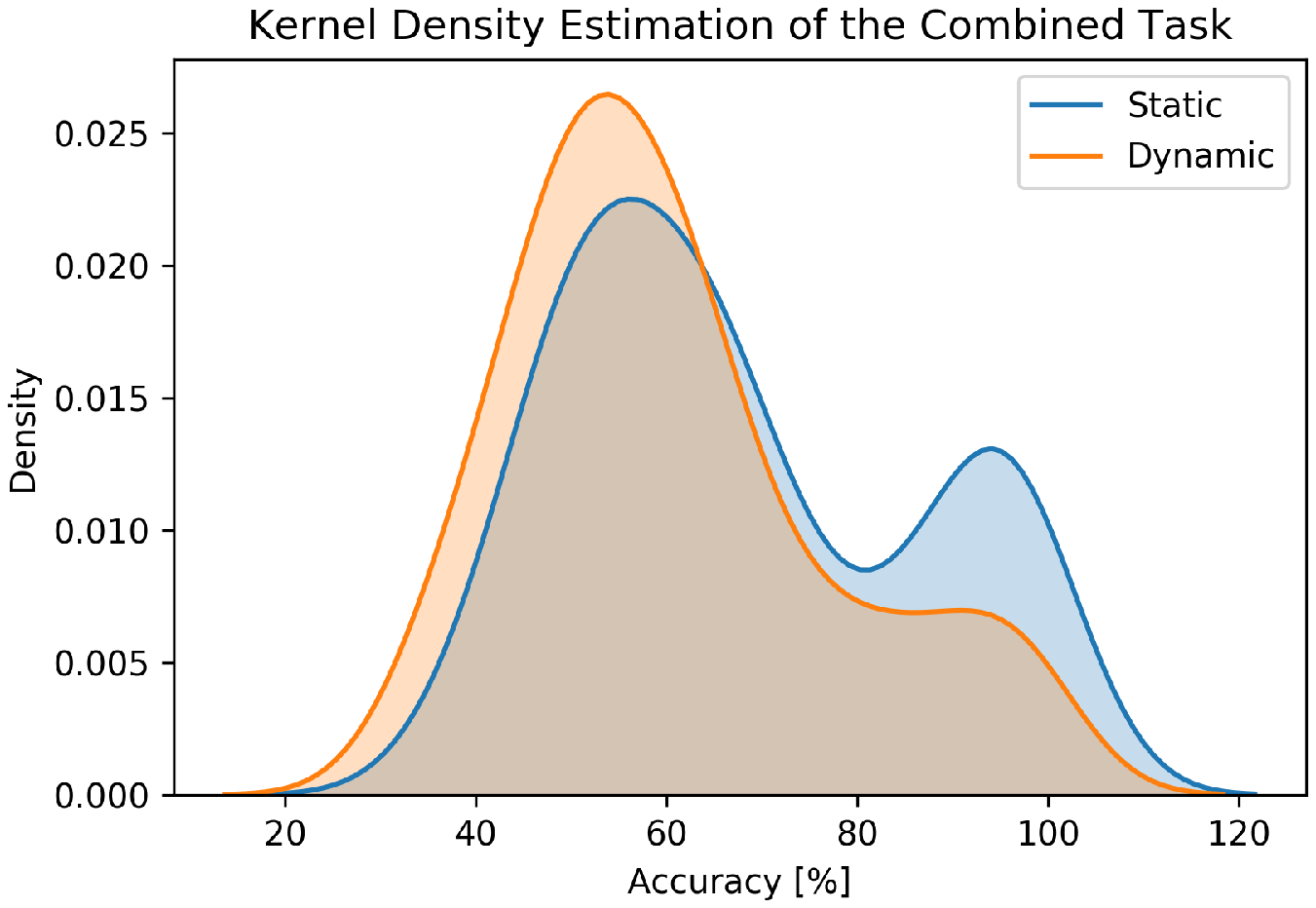}
            \caption{PONO task with reset method}
            \label{reset_combined}
    \end{subfigure}
    \begin{subfigure}[b]{0.33\textwidth}
    \centering
            \includegraphics[width=\textwidth,clip=false]{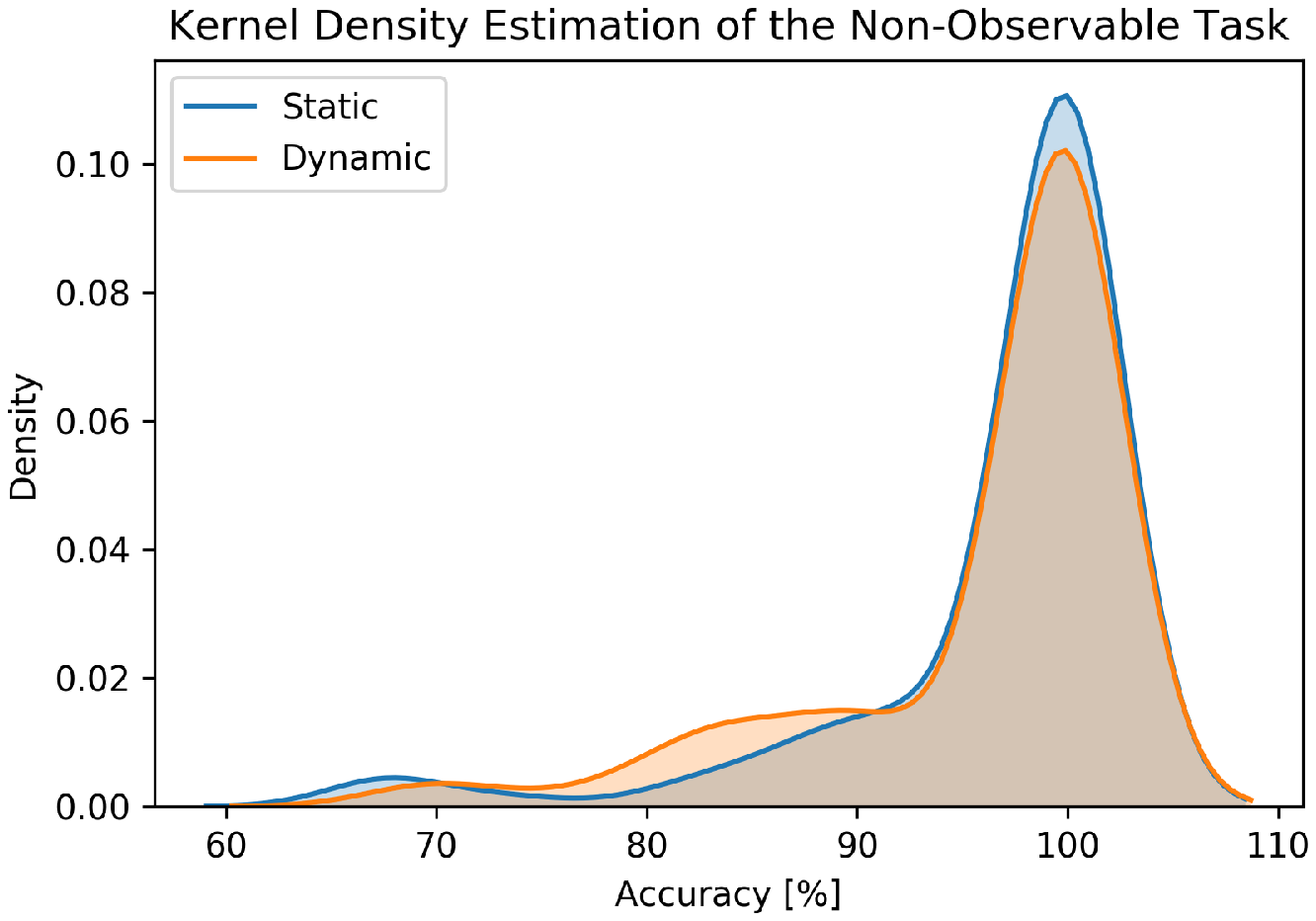}
            \caption{NO task with transfer method}
            \label{transfer_non_obs}
    \end{subfigure}
    \begin{subfigure}[b]{0.33\textwidth}
    \centering
            \includegraphics[width=\textwidth,clip=false]{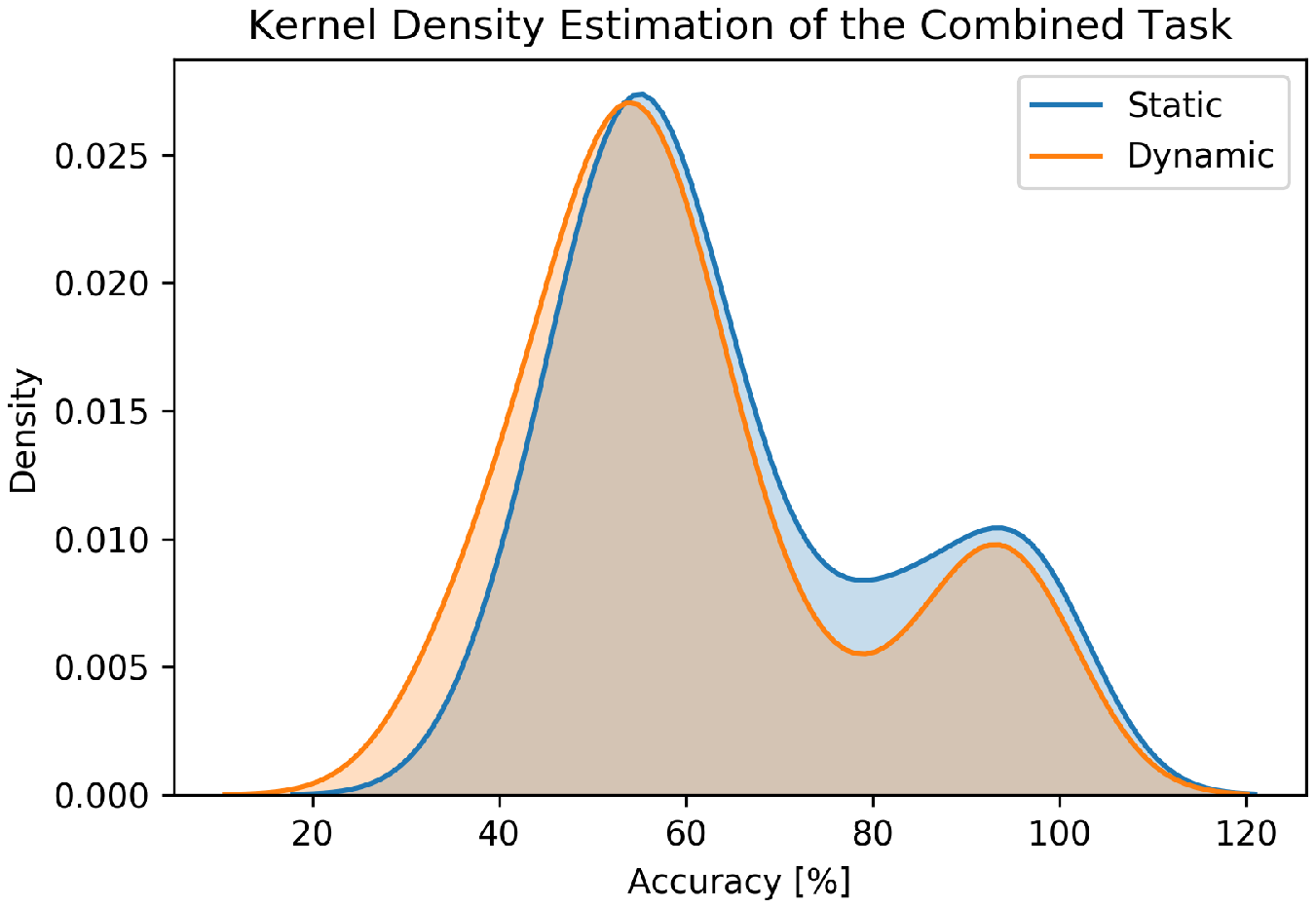}
            \caption{PONO task with transfer method}
            \label{transfer_combined}
    \end{subfigure}
  \caption{The three plots on the first row show the density of the accuracy of the three maze tasks with the reset method, and the bottom show the two relevant tasks for the transfer method.}
\label{fig:KDE}%
\end{figure*}

\subsubsection{Partially-Observable}
To isolate the PO constituent of our model, a maze task using a one-dimensional array with three signals corresponding to three goals was constructed. Since there is only one context, there is no growth in the model, and neither the RM or TM will be utilized. 

At the first time step, the agent is randomly dropped into the maze with the PO feature (signal vector) present in the environment. The agent must learn to use its WM feature to essentially convert this NM task into a M task. As the agent progresses through the task, it is presented with scalar rewards. At states that are not the goal, the agent receives $-1$, and at the goal state, the agent receives $0$. Along with the $0$ reward at the goal state, the agent also receives a goal token, which is just an HRR to identify that a state is the goal. During the training phase, the agent can also use $\epsilon$-soft to explore and escape local minimum. 

\subsubsection{Non-Observable}

To isolate the NO constituent of our model, a maze task with three goals and no signals was constructed. Each of the goals is distributed to one of the three task contexts.  

The agent is randomly dropped into the maze, but there is no signal present. The agent must learn to map the three $atrs$ to the goals. The decision making is similar to the task above, but there are no $wm$ decisions. During the learning and testing phase of the NO task, the context switches after a set number of episodes have passed. The reward system is the same as described above in the PO section.

\subsubsection{Partially-Observable and Non-Observable}
The maze task created for the combined model is more complex than either of the tasks listed above. In this task, there are two abstract tasks with two signals which have different meanings under different contexts.

At time step one, the agent is dropped into the array at a random spot with a PO signal vector present in the environment. Unlike the PO task, the signal might have different meanings depending on the context. The reward system is the same as described above in the PO section.

\section{Results}
The graphs in Figure \ref{fig:KDE} show the results of the model of all relevant tasks in the form of Kernel Density Estimation. The effectiveness of the model is clear for the PO and NO tasks, but the PONO task shows somewhat lower performance due to the lack of tuning the hyper-parameters. 

From \Cref{reset_non_obs}, it is clear the model is able to solve the NO task using the RM with a high rate of success for both static and dynamic thresholds. The dynamic threshold method allows the model to find the optimal threshold, and it is able to perform just as well in the NO task as the static Threshold Method. However, there is still a tail with $60\%-90\%$ accuracy, and this is due to the length of the HRRs and the noise in the hyper-parameters. 

The model performs extremely well on PO tasks especially with the dynamic threshold method, as shown in \Cref{reset_obs}. As with the NO task, there is still a tail, and the way to improve the performance is the same as with the NO task. 

With the PONO task, the model struggles when the testing method does not allow for the tuning of hyper-parameters. Without tuning, \Cref{reset_combined} shows the model to be ineffective with performance as low as $33\%$. After tuning the model, it performs significantly better. Table \ref{params_tuned} shows that the static PONO model is able to improve its performance dramatically by changing the NO task switch rate from 1000 to 2000, and the dynamic PONO model is also able to perform significantly better after tuning. 

There are several ways to tune the hyper-parameters of a model, but we found Bayesian-Optimization (assumes noise is in the hyper-parameters) to be successful. Bayesian-Optimization works by forming a posterior distribution of functions and improving it as observations grow \cite{brochu_bayes_opt}. 

With the TM, as shown in \Cref{transfer_non_obs} and \Cref{transfer_combined}, the model performs well with the retention of information. This means that the model is able to transfer learning from the old values, and continue to take advantage of its previous experiences. This method was not available in the nTL \cite{jovanovich_n_task}, but our model can utilize it with success dramatically reducing overall training time. 

It appears in some cases that the static threshold performs better than the dynamic threshold, but the static threshold cannot learn an arbitrary number of tasks, $n$. Therefore, the dynamic threshold can be seen as more general. 

For near perfect accuracy, the agent needs to use both positive and negative error switching. Table \ref{accuracy_switch} shows the accuracy of the combined model run on the same parameters as tested above but with certain task switching mechanisms removed. As the table shows, it is important for the agent to use both kinds of task switching to achieve near perfect accuracy.

\begin{table}[!t]
\centering
\small
\setlength\tabcolsep{10pt}
\begin{tabular}{c c c} 
 \textbf{Parameters} & \textbf{Static PONO} & \textbf{Dynamic PONO} \\
 \hline
 \hline
 NO Task Switch Rate & 2000 & 2000 \\ 
 ATR Alpha $(\alpha_a)$ & - & 0.000105 \\
 Discount for the trace $(\lambda)$  & - & 0.05 \\
 Accuracy Without Tuning & 38.33\% & 33.57\% \\
 Accuracy With Tuning & 96.88\% & 96.81\% \\
 \hline
\end{tabular}
\vspace{5pt}
\caption{These are the tuned parameters which lead to drastically better performance.}
\label{params_tuned}
\end{table}

\begin{table}[!t]
\centering
\small
\setlength\tabcolsep{7pt}
\begin{tabular}{c c} 
 \textbf{ATR Switching Mechanism} & \textbf{Maximum Accuracy} \\
 \hline
 \hline
 Only Positive Error Switch & 62.56\% \\ 
 Only Negative Error Switch & 48.76\%  \\
 Both Positive and Negative Error Switch & 96.88\% \\
 No Error Switching & 69.45\% \\
 \hline
\end{tabular}
\vspace{5pt}
\caption{This table shows the accuracy of tuned Static PONO from Table \ref{params_tuned} with various task switching mechanisms.}
\label{accuracy_switch}
\end{table}

\section{Discussion}
Since autonomous systems do not perceive the world in the same way as humans, they need to form their own internal representations. The agent must be able to disregard irrelevant information from the environment and use relevant information to solve the current task. To give autonomous systems the ability to do this, the HWMtk was created \cite{dubois_hrr} inspired by the putative interactions of the PFC and BG for WM and higher-order cognition. For tasks where the environment does not have all relevant information, nTL was created \cite{jovanovich_n_task}. 

These two distinct models described can either utilize ATRs or storage of past input features but not both. However, in the real world, tasks are not always so simple: they may contain both PO and NO features. In this paper, we presented a model, PONOWMtk, that can utilize both methods to solve tasks with both PO and NO features. Furthermore, the model can solve both a static and a dynamic number of tasks with the use of a dynamic threshold, and the model can use the TM to retain the value function while also growing. 

In the future, we would like to continue to explore TM and more complex neural networks. The model also implicates the PFC and BG as key constitutes to higher-order cognition involved in the solution of PO, NO, and PONO tasks. 

\bibliographystyle{unsrt}  
\bibliography{khan}  

\begin{thebibliography}{10}

\bibitem{Fukuda_how_far_away_is_man}
T.~{Fukuda}, R.~{Michelini}, V.~{Potkonjak}, S.~{Tzafestas}, K.~{Valavanis},
  and M.~{Vukobratovic}.
\newblock How far away is "artificial man".
\newblock {\em IEEE Robotics Automation Magazine}, 8(1):66--73, March 2001.

\bibitem{Tugcu_neuroscience_model_of_working_memory}
Mert Tugcu, Xiaochun Wang, Jonathan~E Hunter, J~Phillips, D~Noelle, and
  Don~Mitch Wilkes.
\newblock A computational neuroscience model of working memory with application
  to robot perceptual learning.
\newblock In {\em Third IASTED International Conference on Computational
  Intelligence (CI)}, 2007.

\bibitem{jovanovich_n_task}
Michael Jovanovich and Joshua Phillips.
\newblock n-task learning: Solving multiple or unknown numbers of reinforcement
  learning problems.
\newblock In {\em CogSci}, pages 584--589, 2018.

\bibitem{phillips_wmtk}
Joshua~L Phillips and David~C Noelle.
\newblock A biologically inspired working memory framework for robots.
\newblock In {\em ROMAN 2005. IEEE International Workshop on Robot and Human
  Interactive Communication, 2005.}, pages 599--604. IEEE, 2005.

\bibitem{dubois_hrr}
Grayson~M DuBois and Joshua~L Phillips.
\newblock Working memory concept encoding using holographic reduced
  representations.
\newblock In {\em MAICS}, pages 137--144, 2017.

\bibitem{Reilly_neuromodulatory}
Randall~C O'Reilly, David~C Noelle, Todd~S Braver, and Jonathan~D Cohen.
\newblock Prefrontal cortex and dynamic categorization tasks: representational
  organization and neuromodulatory control.
\newblock {\em Cerebral cortex}, 12(3):246--257, 2002.

\bibitem{collins_neurogenetic}
Anne~GE Collins and Michael~J Frank.
\newblock How much of reinforcement learning is working memory, not
  reinforcement learning? a behavioral, computational, and neurogenetic
  analysis.
\newblock {\em European Journal of Neuroscience}, 35(7):1024--1035, 2012.

\bibitem{baddeley_working_memory}
Alan Baddeley.
\newblock Working memory.
\newblock {\em Science}, 255(5044):556--559, 1992.

\bibitem{kunz_td}
Florian Kunz.
\newblock An introduction to temporal difference learning.
\newblock In {\em Seminar on Autonomous Learning Systems}, 2000.

\bibitem{sutton_RL_intro}
Richard~S Sutton and Andrew~G Barto.
\newblock {\em Reinforcement learning: An introduction}.
\newblock MIT press, 2018.

\bibitem{plate_holographic}
Tony~A Plate.
\newblock Holographic reduced representations.
\newblock {\em IEEE Transactions on Neural networks}, 6(3):623--641, 1995.

\bibitem{Tange_parallel}
O.~Tange.
\newblock Gnu parallel - the command-line power tool.
\newblock {\em ;login: The USENIX Magazine}, 36(1):42--47, Feb 2011.

\bibitem{brochu_bayes_opt}
Eric Brochu, Vlad~M Cora, and Nando De~Freitas.
\newblock A tutorial on bayesian optimization of expensive cost functions, with
  application to active user modeling and hierarchical reinforcement learning.
\newblock {\em arXiv preprint arXiv:1012.2599}, 2010.

\end{thebibliography}

\end{document}